\documentclass[10pt,twocolumn,letterpaper]{article}

 \usepackage{cvpr}              

\usepackage{graphicx}
\usepackage{amsmath}
\usepackage{amssymb}
\usepackage{booktabs}
\usepackage{paralist}
\usepackage{multirow}

\usepackage[pagebackref,breaklinks,colorlinks]{hyperref}

\usepackage[capitalize]{cleveref}
\crefname{section}{Sec.}{Secs.}
\Crefname{section}{Section}{Sections}
\Crefname{table}{Table}{Tables}
\crefname{table}{Tab.}{Tabs.}

\def\cvprPaperID{5355} 
\def\confName{CVPR}
\def\confYear{2022}

\begin{document}

\title{ Finding Fallen Objects Via \\Asynchronous Audio-Visual Integration}

\author{Chuang Gan\thanks{Equal Contribution}, Yi Gu$^{\ast}$, Siyuan Zhou, Jeremy Schwartz, Seth Alter, James Traer, \\ Dan Gutfreund, Joshua B. Tenenbaum, Josh H. McDermott$^{\ast}$, Antonio Torralba$^{\ast}$\\
 MIT and MIT-IBM Watson AI Lab}

\maketitle

\begin{abstract}
The way an object looks and sounds provide complementary reflections of its physical properties. In many settings cues from vision and audition arrive asynchronously but must be integrated, as when we hear an object dropped on the floor and then must find it. In this paper, we introduce a setting in which to study multi-modal object localization in 3D virtual environments. An object is dropped somewhere in a room. An embodied robot agent, equipped with camera and microphone, must determine what object has been dropped -- and where -- by combining audio and visual signals with knowledge of the underlying physics. To study this problem, we have generated a large-scale dataset -- the Fallen Objects dataset -- that includes 8000 instances of 30 physical object categories in 64 rooms. The dataset uses the ThreeDWorld Platform that can simulate physics-based impact sounds and complex physical interactions between objects in a photorealistic setting. As a first step toward addressing this challenge, we develop a set of embodied agent baselines, based on imitation learning, reinforcement learning, and modular planning, and perform an in-depth analysis of the challenge of this new task. This dataset is publicly available \footnote{Project page: \url{http://fallen-object.csail.mit.edu}}.
   
\end{abstract}

\section{Introduction}
\label{sec:intro}

Humans integrate multi-sensory data to understand the physical world around us. Consider the situation in which we hear the sound of an object falling somewhere in our house. What was it that fell, and where? Just by listening to the sound that is produced, we can usually determine not only the approximate location of the object that fell but also aspects of its physical make-up. The loudness of the sound, along with the reverberation that accompanies it, tells us much about the object's size, force of impact, and distance~\cite{traer2021envsound}. And we can usually tell whether the object was metal, wood or plastic~\cite{giordano2006material, traer2019impacts}, and whether it rolled on the floor after impact~\cite{agarwal2021scrapingrolling}. However, the sound alone is often not enough to precisely reveal the identity and location of the object, and instead must be used to guide visual search. Once we are in the vicinity of the fallen object, we use vision to find the object on the floor that is consistent with what we heard. We term this `asynchronous audio-visual integration'.

\begin{figure}[t]
    \centering
  \includegraphics[width=\linewidth]{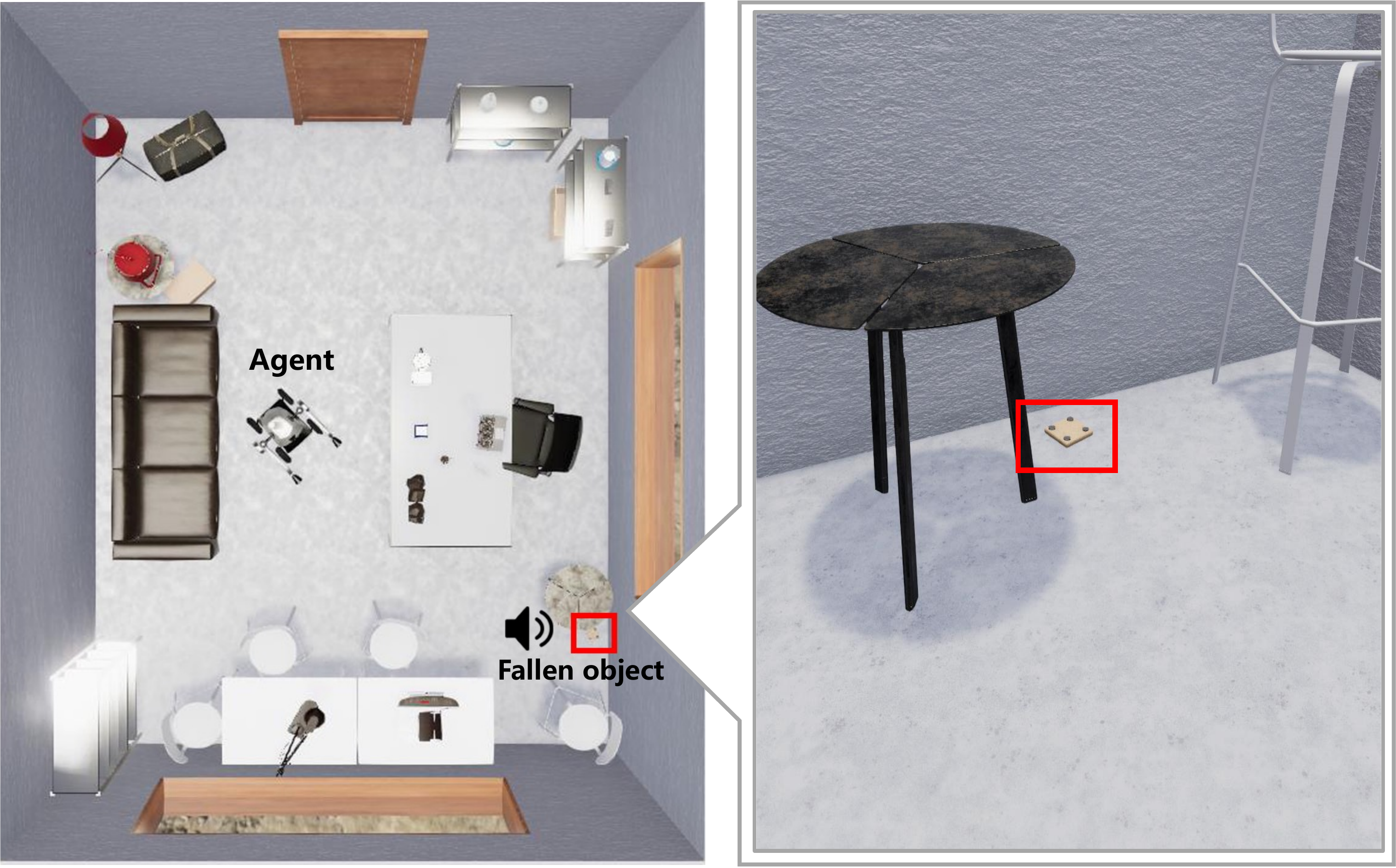}
    \caption{Illustration of the proposed embodied physical sound source localization: an agent hears a physical object fall somewhere in the same room, and is required to find it via asynchronous audio-visual integration.}
    \label{fig:teaser}
\end{figure}

Replicating similar capabilities in assistive robots will be useful for many real-world applications. For example, robots may need to fetch a screw dropped in the middle of an automated production line which might shut down operations until the item is located and safely removed. The field of audio-visual navigation has made exciting progress by using both visual and audio modalities to drive embodied agents to navigate towards the target location of sound sources. Some of this earlier work~\cite{gan2020look,chen2020soundspaces} defines the acoustic target as a repeating sound, such as a phone ringing or an alarm, providing a constant acoustic cue to the agent. However, in real-world situations, many sounds are intermittent, or are the result of a single non-repeating impact event. Recent work from Chen~\cite{chen2021semantic} made the first attempt to introduce semantic audio-visual navigation, in which objects in the environment must be localized via sound clips of short duration. The use of brief, non-repeating sounds was an advance over previous work, but the resulting dataset had two limitations. First, the platform that was used does not include impact sound synthesis capability, and therefore was unable to render the physical properties and interactions of objects via audio. The audio that was used did not reflect any underlying physical events such as objects falling, bouncing and possibly colliding with other objects, nor did it reflect the physical parameters of objects. As a result, the task could plausibly be solved primarily by learning semantic correlations between sounds and the scene context. Second, the locations of the sounding sources were defined in 2D uniform grids with fixed height, rather than the 3D locations that characterize sound source localization in real-world environments (in which the sounding objects could be anywhere in the room).

To remedy these limitations, we propose a new embodied AI challenge for multi-modal physical scene understanding. An embodied agent hears an unknown object fall to the ground, somewhere in the room it is in. The agent must then navigate within the room to locate and identify this physical object using both visual and auditory modalities (Figure~\ref{fig:teaser}). To support this task, we use the ThreeDWorld (TDW) simulation platform~\cite{gan2020threedworld}, which provides real-time synthesis of impact sounds, photo-realistic rendering and believable physical simulation. Audio is rendered with TDW’s PyImpact Python library, which uses modal synthesis to generate audio from information about the material types and impact parameters of colliding objects (velocities, normal vectors and masses)~\cite{traer2019impacts}. Impact sounds are then spatialized using Resonance Audio, providing reverberation simulation that reflects the spatial dimensions of the room and the wall and floor materials being used. 

We incorporated physical simulation of objects' physical interaction behavior and resulting impact sounds into embodied audio-visual navigation so as to pose several unique challenges for the agent. First, the diversity of locations of the target objects is increased relative to previous work, since the objects may be under or behind a sofa, on top of a cabinet, on a shelf, inside another containing object, or under a table. In particular, the agent is required to adjust its height in order to find the fallen object successfully. Second, we included distractor objects on floors, tabletops, and counter surfaces that disguise the location of target objects. Thus, the agent cannot simply leverage the correlation of the sound and scene to navigate towards the object. Instead, it must use the audio signal to infer the physical properties of the fallen object, constraining what this fallen object looks like and roughly where it has fallen.

We evaluate several agents on this benchmark. Though recent progress in embodied navigation has resulted in agents capable of navigating within an unfamiliar environment, experimental results suggest that it remains very challenging for these embodied agents to complete our task successfully. We believe models that perform well on our challenge will take a meaningful step towards more intelligent robots that could infer physical information about the scene from multi-sensory data. Our contributions are summarized as follows:
\setdefaultleftmargin{1em}{1em}{}{}{}{}
\begin{compactitem}
  \item We introduce a new \textit{embodied physical sound source localization} task that aims to measure AI agents' physical inference abilities by finding fallen objects via asynchronous audio-visual integration.

 \item To support this challenge, we augment the existing TDW simulation platform by adding 64 new physical rooms and 8 new audio materials to simulate a diverse range of impact sounds in the different scenes.

  \item We create the \textit{Fallen Objects} dataset, a comprehensive audio-visual dataset that includes over 8000 instances of 30 physical object categories fallen in 64 rooms. We will make this dataset publicly available. 
  
  \item We develop several baseline agents as first steps to tackle this task. We use these baselines to perform in-depth analysis of the challenges presented by our benchmark, and also highlight potential directions to improve performances on the task. 
    
\end{compactitem}

\section{Related Work}
\label{sec:related_work}

\vspace{2mm}
\noindent\textbf{Embodied AI.}
A long-standing goal for the computer vision and robotics communities is to develop intelligent agents that could assist with everyday tasks. Since directly training robot agents in the real world is expensive, and makes it hard to benchmark different algorithms, there has been growing interest in using high-fidelity interactive 3D simulation to train and evaluate embodied agents. Representative platforms include Habitat~\cite{savva2019habitat}, Gibson~\cite{xia2018gibson}, i-Gibson~\cite{li2021igibson}, AI2-Thor~\cite{kolve2017ai2}, Virtual Home~\cite{puig2018virtualhome}, Sapien~\cite{xiang2020sapien}, and ThreeDWorld~\cite{gan2020threedworld}. These platform have also been used to generate large-scale datasets, making progress on pointing goal navigation~\cite{savva2019habitat,wijmans2019dd}, semantic navigation~\cite{chaplot2020object}, interactive navigation~\cite{li2021igibson}, instruction following~\cite{shridhar2020alfred}, audio-visual navigation~\cite{chen2020soundspaces,gan2020look}, rearrangement~\cite{gan2021threedworld,batra2020rearrangement,RoomR,srivastava2021behavior}, among other tasks.   State-of-the-art approaches to embodied tasks include end-to-end training of neural policies using RL~\cite{zhu2017target, wijmans2019dd} or hierarchical RL~\cite{kaufmann2019beauty,bansal2020combining} and modular approaches by integrating perception and mapping for path planning~\cite{gupta2017cognitive,chaplot2020learning,chaplot2020object}. Recent work has also shown that web videos can be used to train embodied agents that could flexibly navigate to a goal~\cite{chang2020semantic,hahn2021no}.  However, at present we lack datasets with which build multi-modal physical reasoning abilities into embodied agents. Our challenge aims to fill this gap.

\vspace{2mm}
\noindent\textbf{Audio-Visual Navigation.}
Audio-visual navigation has recently emerged as a research focus in computer vision and robotics~\cite{gan2020look,chen2020learning,chen2020soundspaces,chen2021semantic,dean2020see}. Unlike visual navigation, an audio sound source is used to define the target, and the agent must use both audio and visual signals to navigate to the target. Existing platforms used for this task simulate audio using a prerecorded repeating sound (\eg alarm)~\cite{gan2020look,chen2020soundspaces} or a brief semantically identifiable sound~\cite{chen2021semantic} downloaded from the Internet. Then they use acoustic reverberation simulations to calculate the impulse response at each of several pairs of sound emitter and receiver locations. Finally, they convolve the impulse response with the raw waveform to simulate the sound observation for the agent. As discussed above, this simplification is a useful first step towards incorporating audio signals into embodied AI, but it does not consider how the physical properties of objects, and the physics of their interactions, influence what the agent hears. By contrast, human listeners are believed to implicitly infer physical generative parameters from what they hear, and to use these inferences in estimating the distance of sound sources~\cite{traer2021envsound}. In this work, we remedy these limitations by incorporating new simulation techniques of physically-driven impact sound synthesis for complex object interactions in near photo-realistic environments. These enable us to create a large-scale audio-visual dataset of fallen objects that poses unique challenges for object-centric physical inference from multi-sensory data.

\vspace{2mm}
\noindent\textbf{Audio-Visual Learning.}
Our work is also related to a body of research leveraging the relationship between vision and sound for model learning. A main theme of this previous work has been to show that the synchronization of audio and video can be utilized as a `free' training signal. The applications include feature representation learning~\cite{arandjelovic2017look,owens2018audio,owens2016ambient}, visually-guided sound separation and localization~\cite{zhao2018sound,gao2018learning,ephrat2018looking}, object localization~\cite{arandjelovic2018objects,gan2019self,afouras2020self}, scene parsing~\cite{tian2020unified,rouditchenko2019self}, active speaker detection~\cite{roth2020ava}, depth prediction~\cite{gao2020visualechoes}, learning of scene structure~\cite{chen2021structure}, floor plan reconstruction~\cite{purushwalkam2021audio}, cross-modal generation\cite{owens2016visually,gao20192,oh2019speech2face,shlizerman2018audio,su2020audeo,gan2020foley} and so on. Our work differs from these prior efforts in studying how to use asynchronous visual and audio information to perform physics-based tasks.

\begin{figure*}[t]
    \centering
  \includegraphics[width=\linewidth]{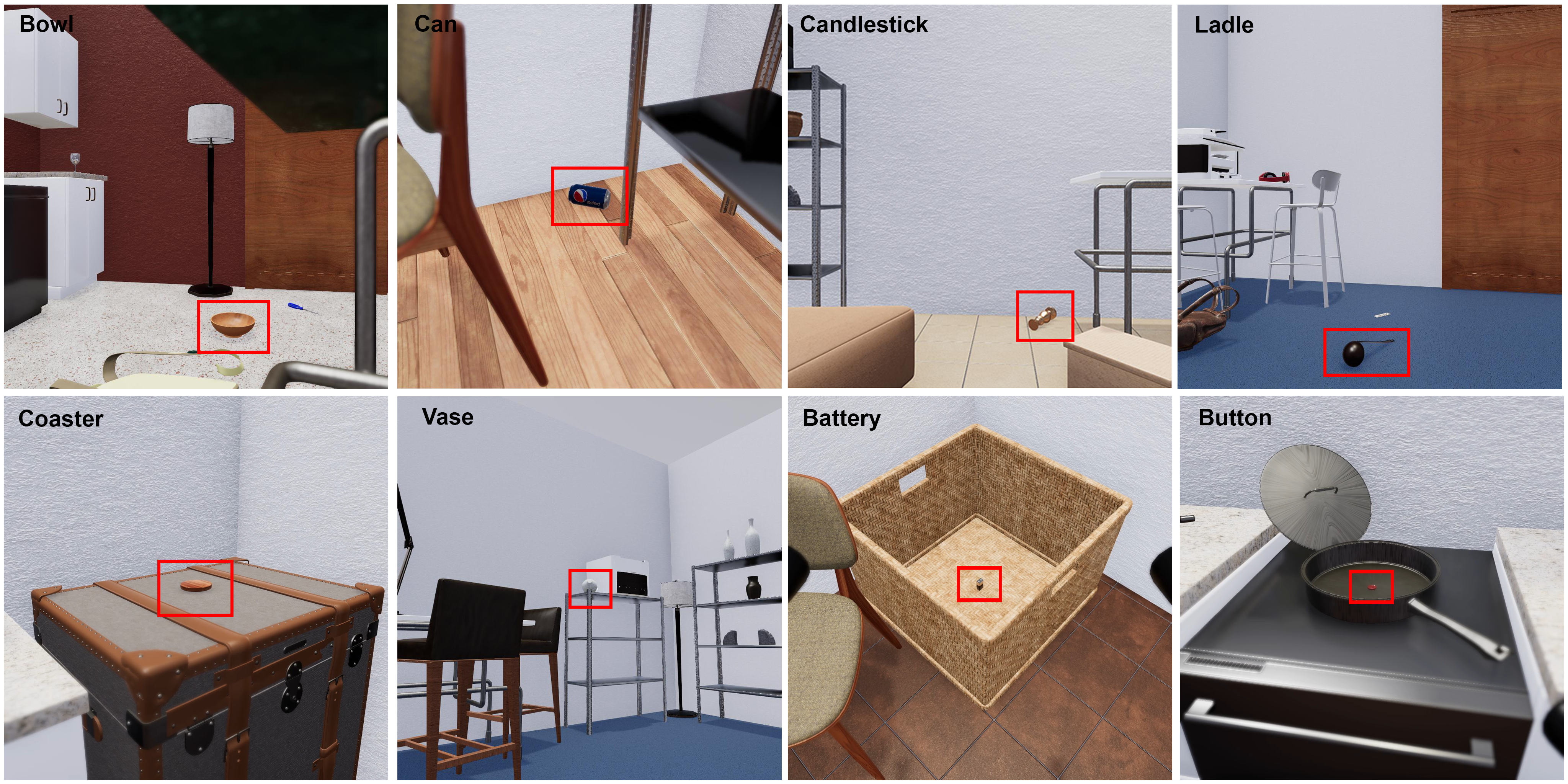}
    \caption{An overview of the fallen objects dataset.  The objects and fall locations are diverse; objects like those depicted in the images can land on the floor, on a table or trunk, on top of a shelf, or hidden within containers such as a basket or a pan on the stove.}
    \label{fig:dataset}
\end{figure*}

\section{Fallen Objects Dataset}
\label{sec:dataset}

We built the \textit{Fallen Objects} dataset to assess an agent's asynchronous audio-visual integration capabilities for physical object localization in simulated 3D environments that mimic the sensory and interactive richness of the real world. We chose to use the ThreeDWorld (TDW) platform~\cite{gan2020threedworld} to generate the dataset because TDW supports both near-photo-realistic image rendering, physically-based sound rendering~\cite{traer2019impacts} with reverberation, and realistic physical interactions between objects and agents. In this section, we introduce the problem formulation and data collection process. 

\subsection{Problem Formulation}

\vspace{2mm}

\noindent\textbf{Task definition.}
An embodied agent with an egocentric-view camera and microphone hears an unknown object fall somewhere in the same room it is in (a study or kitchen room). The agent is then required to find which object has fallen and where, as efficiently as possible, using asynchronous audio-visual integration. 

\vspace{2mm}

\noindent\textbf{Observation and action space}
The observation space for this agent contains a first-person-view RGB image, depth map, egocentric binaural audio as well as the agent's current pose. The audio observation is only available at the very beginning, when the object falls and settles (\ie step 0). The agent's action space comprises \textit{move forward, rotate left, rotate right, look up, look down}, and \textit{found}. 

\vspace{2mm}
\noindent\textbf{Success criterion.}
The agent succeeds on the task if and only if it positions itself within 2 meters of the target object~\cite{chang2020semantic}, with the target object appearing within the first-person view when the agent executes the \textit{found} command. This is a strict success criterion inspired by consideration of real-world applications, in which an assistive robot needs to explicitly understand the goal and whether it has reached it. 

\subsection{Dataset Generation}
Simulating this type of large-scale physical object event data involves several challenges. First, it requires a simulation platform that can deeply integrate physics behavior, impact-based sound synthesis, and image rendering, while also supporting an embodied agent that can interact with the virtual world. Second, it requires a rich set of object assets and material types that can generate a comprehensive and diverse set of physical object falling events. We chose the ThreeDworld platform because its advanced multi-modal data synthesis, physical simulation capabilities, and physical object library provide solutions to these challenges. 

\vspace{2mm}
\noindent\textbf{Simulating fallen physical objects in 3D environments.} To support the dataset generation, we created 8 new 3D models of kitchen and study room environments, to be used in the TDW simulation engine. For each room environment we varied the wall and floor materials and included one of 4 different furniture layouts, for a total of 64 room variants. All objects in these room environment scenes respond to physics. We also define a set of 30 object categories that will be dropped under the control of physics; these objects are assigned varying physical material properties such as mass, friction and restitution (bounciness). Objects are additionally assigned an audio material that matches their variation in visual material (e.g. wood, metal, ceramic etc.) For this challenge, we created 8 new audio material types, including both hard and soft plastics, rubber, stone and several additional wood materials, for a total of 14 materials.  When objects collide with other objects or the floor, TDW's PyImpact module synthesizes impact sound data reflecting the object's physical and material properties as well as parameters of the collision such as velocity and angle of impact. Impact sounds are spatialized, providing real-time audio propagation, directional cues via head-related transfer function, and high-quality simulated reverberation that varies with room geometry and surface materials. Target object categories include: keys, coasters, cutlery, glasses, batteries, coins and corks, each using one of the 14 materials. The full list can be found in the supplementary material. 

\vspace{2mm}
\noindent\textbf{Procedural generation of impact sounds.}
For each room variant, we define 20 \textit{fall zones}, i.e. locations where we expect the target objects to land when fallen.  Fall zones can be on the floor, on top of tables, shelves  or cabinets, or within open container objects such as boxes, baskets or pans on the stove. Fall zones can also contain additional ``distractor" objects of similar size to the target object, as well as occluder objects that can hide it, both of which increase the difficulty of finding the target object. Some examples of fallen objects can be found in Figure~\ref{fig:dataset}.  

To efficiently generate valid scenarios for this dataset, we developed a two-phase procedural generation pipeline: an initial rehearsal phase that generates guaranteed-valid fall event data, and a subsequent generation phase that uses that data to generate audio heard by the agent. Thus, before rendering any audio data, we removed any failed trial scenarios where objects fell outside of the desired fall zone.
\vspace{-2mm}
\begin{enumerate}
\item \textbf{Rehearsal Phase.} We simulate target objects dropping onto their designated fall zone. Objects that fall or roll outside of their fall zone, and end up in plain view, are considered too easy for the agent to find and are filtered out. For each room variant, we initialize the scene and fall zone data. We then run a series of trials that: 1) define a randomized starting position and orientation above the fall zone for each target object; 2) drop the object under control of physics. All objects that remain within the drop zone when they cease moving are kept for the dataset. We record the start and end parameters for these objects (termed the `fall event data') for the generation phase, described next.

\item \textbf{Generation Phase.}
In this phase, we generate the actual audio produced by the fallen objects. For each room variant, we load the corresponding fall event data from the rehearsal phase, and initialize the reverberation and wall/floor material parameters. We then generate a series of trials, where each trial 1) spawns an embodied agent in a random free position and facing in a random direction; 2) adds the target object from the fall event and lets it fall; 3) generates a spatialized audio recording of the falling event, as heard by the agent, using the TDW sound engine. 
\end{enumerate}

\vspace{2mm}
\noindent\textbf{Collecting demonstrations data of expert trajectories.}
We further collect demonstration data for all scenarios to illustrate the shortest path and optimal action sequences to succeed in the task. We first calculate a shortest path to navigate from the agent's position to the fallen object based on the ground truth occupancy maps from the the simulation. After the agent arrives near the fallen object, we then adjust its orientation by executing actions \textit{rotate left} or \textit{rotate right} and its height by executing actions \textit{look up} or \textit{look down} to search for the target object. These expert trajectory demonstration data are intended to help researchers improve policy learning or incorporate heuristics into planning algorithms. 

\section{Experiments}

\subsection{Baselines}
We implemented 5 baseline agents that include models using imitation learning, reinforcement learning and modular planning-based algorithms:

\vspace{2mm}
\begin{compactitem}

\item \textbf{Decision Transformer~\cite{chen2021decision}:} This is a recent state-of-the-art imitation learning framework, which leverages the transformer architecture~\cite{vaswani2017attention} to model trajectories and predict future actions.  There are 3 kinds of embeddings: state, action and return-to-go. We extract the features from RGBD images, sound spectrograms and segmentation masks as state embeddings and learn a linear layer for action and return-to-go embeddings. In addition, we learn timestep embeddings and add them to each token. We finally feed the last 20 timesteps tokens into the model and use a GPT model~\cite{radford2018improving} to predict future actions. The model is trained using the expert trajectories provided as part of our dataset.

\item \textbf{PPO~\cite{schulman2017proximal}:} 
Similar to previous work~\cite{chen2020soundspaces}, we train an end-to-end RL policy using Proximal Policy Optimization (PPO) by maximizing the reward of finding the audio goal (fallen objects in our case). This model takes the input history of RGBD images, segmentation masks as well as audio observations and outputs actions that the agent should execute in the environment. We adopt an oracle \textit{found} action for this agent.  Specifically, the agent will be considered successful if the fallen object is less than 2 meters from the agent while also appearing in the first-person-view image.

\item \textbf{SAVi~\cite{chen2021semantic}:} This is a recent state-of-the-art framework for semantic audio-visual navigation. It first uses an auditory perception module to estimate an audio goal and then adopts a transformer-based deep RL framework that can learn to attend to the previous visual and audio observations to determine a path to the predicted audio goal. We use their open-source code to perform our task. 

\item \textbf{Modular Planning~\cite{chaplot2020object,gan2020look}:} This is a hybrid and modular framework that integrates the auditory perception module used in~\cite{gan2020look} and the visual semantic mapping module used in~\cite{chaplot2020object}. The model first predicts the fallen object's category and location based on the audio, and then builds maps of the environment from RGBD images in search of the fallen object. If the agent does not find the goal object successfully during this process, it will adopt a frontier exploration strategy~\cite{FBE} to search unexplored areas. Otherwise, it will move close to the potential fallen object and find it.

\item \textbf{Object-Goal~\cite{chaplot2020object}:} This is a recent state-of-the-art method for visual semantic navigation. It adopts a modular SLAM-based navigation approach. They first use depth projection and segmentation for semantic mapping and select a waypoint to explore. If the target object appears in the map, the agent plans a shortest path and navigates towards it. We use this baseline to test whether or not the agent needs audio signals to succeed on our task.  To decide when to execute the \textit{found} action, we provide 5 potential object categories to the agent (one of which is correct), which uses a heuristic approach to guess if an object is fallen or not. We used the open-source code from the original paper~\cite{chaplot2020object} and adapted it to our task. 

\end{compactitem}

\subsection{Experimental Setup}

\vspace{2mm}
\noindent\textbf{Setup.} We use 8000 instances of 30 physical object categories in 64 physically distinct rooms (32 study rooms and 32 kitchens) for the experiments. We used 6000 of these instances for training, 1000 for validation, and 1000 for testing. We report the models' performance on the test set. The agent received an audio observation only at the start of the trial. The agent then received an RGB-D observation at each subsequent time step. For the actions, we set \textit{move forward} to 0.25m and \textit{rotate} to 30 degrees. To test the generalization capabilities of these agents, we additionally carried out cross-scene generalization experiments described in section~\ref{sec:cross}. 

\vspace{1mm}
\noindent\textbf{Evaluation metrics.} We use three metrics to evaluate agents: Success Rate, Success weighted by Path Length (SPL)~\cite{anderson2018evaluation}, and Success weighted by Number of Actions (SNA)~\cite{chen2020learning}.  

The \textit{Success Rate} is defined as the ratio of the number of times the agent successfully navigates to the target to the total number of test trials. A trial is considered successful based on four criteria: first, the agent needs to explicitly execute action \textit{found}; second, the distance between the agent and the goal is less than 2 meter; third, the target physical object should appear in the agent's view; fourth, the length of the action sequence is less than the maximum number of allowed time steps $N$. We set $N$ as 200 for all comparisons, as this was twice the average action sequence length of the demonstration data.  The \textit{SPL} is a metric that jointly considers the success rate weighted by the path length to reach the goal from the starting point. And the \textit{SNA} jointly considers the number of actions and the success rate, by penalizing collisions, rotations, and height adjustments needed to find the objects.

\subsection{Implementation Details}
For all the baselines, the input observation data are 300 $\times$ 300 size RGBD images, around 2 second binaural audio sampled at 44.1 kHz , and agent pose.  Below we provide the details of the baseline models: the visual and auditory perception modules, semantic goal and occupancy map estimation, policy training, and the planners.

\vspace{2mm}
\noindent\textbf{Visual and Auditory Perception Modules}
We first train a Mask-RCNN model~\cite{he2017mask} as a visual perception module for semantic segmentation of 30 pre-defined object categories. We then train an auditory perception module used for anticipating the category and location of the sounding fallen object in our environments. To pre-process the audio data, we transform the raw waveform into a spectrogram, which provides the input to a ResNet-18 network~\cite{he2016deep}. The network is trained to predict the object category from sound using a cross-entropy loss. To estimate the location of the fallen object, we follow previous work~\cite{gan2020look} by predicting the relative orientation and distances with respect to the agent, trained using a mean squared error (MSE) loss. The absolute coordinate of the estimated fallen object can then be calculated from the agent's coordinates and the relative location. 

\vspace{2mm}
\noindent\textbf{Semantic Goal and Occupancy Map.}
We represent the global top-down semantic goal map as $S_g \in \{0, 1\}^{N \times N}$ and the occupancy map as $ O_g \in \{0, 1\}^{2 \times N \times N}$, where N$\times$N represents the map size and each cell corresponds to a region of size 0.1m$\times$0.1m in the room. Each cell in the semantic goal map represents whether the corresponding region of the room contains a possible target object or not. For a region that might contain a possible target object, we map the geometric center of this semantic segmentation mask to a cell using the depth map.  The two channels in the occupancy map represent whether the cell is occupied and explored, respectively. This is measured by whether the corresponding region contains an obstacle or has been observed.
At each time step, the agent can recover a local semantic map $S_l \in \{0, 1\}^S$ and a local occupancy map $ O_l \in \{0, 1\}^{2 \times S}$ from the egocentric semantic segmentation and depth map of the visible area in front of the agent.

\vspace{1mm}
\noindent\textbf{Planning.} For planning, the agents first rely on the auditory perception module to infer a goal location and potential fallen object category from the sound. As the agents receive RGBD images at each time step, we run a visual perception module on the RGB images, returning segmentation masks of objects in the egocentric view. We then integrate the semantic segmentation with the depth image to construct an occupancy map and a semantic goal map for path planning. The planning module needs to correlate the semantic goal map with the object category inferred from the auditory perception module. The planner will provide a sequence of actions to move towards this target object along the shortest path if the target object is found around the position estimated from the audio. If the sound position prediction does not correspond to anything in the semantic goal map, the agent will use frontier exploration to navigate the unexplored area on the occupancy map in search of the fallen object. 

\vspace{2mm}
\noindent\textbf{Policy Training.}
For training PPO and SAVi agents, we use the same reward function for policy training. At each step, the agent receives a reward of $+1$ if it is close to the target location and $-1$ if it is far from it.  The agent receives a reward of $+10$ if it finds the fallen object. There is also a $-0.01$ penalty for each time step.

For PPO, we first use CNNs to encode the RGBD images, the semantic segmentation and the spectrogram at each time step as feature vector and concatenate them as an input for the Gated Recurrent Unit (GRU) model. We train PPO with the Adam optimizer, using a learning rate of $2.5 \times 10^{-4}$. We collect the experience of $16$ steps to update the policy. For SAVi, we use the same auditory perception module to initialize the weight of the target descriptor. The location prediction network is then fine-tuned with the ground truth locations online during policy training. We train SAVi with Adam using a learning rate of $2.5 \times 10^{-4}$. Follow the implementation in~\cite{chen2021semantic}, we also roll out the policy for $150$ steps and then use these experience data to update the policy every two epochs. We train both the PPO and SAVi models until convergence. 

\begin{table*}[!h]
\centering
\caption{Navigation performance comparisons on the fallen object dataset. The modular planning-based agent achieves the best results. We also provide additional diagnosis results for this agent by replacing each individual component as an oracle module.}
\begin{tabular}{l|cccc}
   \hline
Method & Success Rate & SPL & SNA  \\
 \hline
 Decision TransFormer~\cite{chen2021decision} & 0.17 & 0.12 & 0.14 \\
 PPO  (Oracle found)~\cite{schulman2017proximal}  &  0.19 & 0.15 & 0.14 \\
  SAVi~\cite{chen2021semantic}   &  0.23 & 0.16 & 0.10 \\
Object-Goal~\cite{chaplot2020object}  &  0.22 &  0.18 & 0.17 \\
Modular Planning~\cite{gan2020look}  & \textbf{0.41} & \textbf{0.27} & \textbf{0.25} \\
\hline
Modular Planning (GT seg.)  & 0.47 & 0.34 & 0.32 \\
Modular Planning (GT object)  & 0.67 & 0.41 & 0.38  \\
Modular Planning (GT seg.+ object)  & 0.85 & 0.61 & 0.56  \\
Modular Planning (GT seg.+ object + location)  & 0.93 & 0.83 & 0.80 \\

\hline
\end{tabular}
    \label{tab:main}
\end{table*}

\begin{figure*}[!h]
    \centering
  \includegraphics[width=\linewidth]{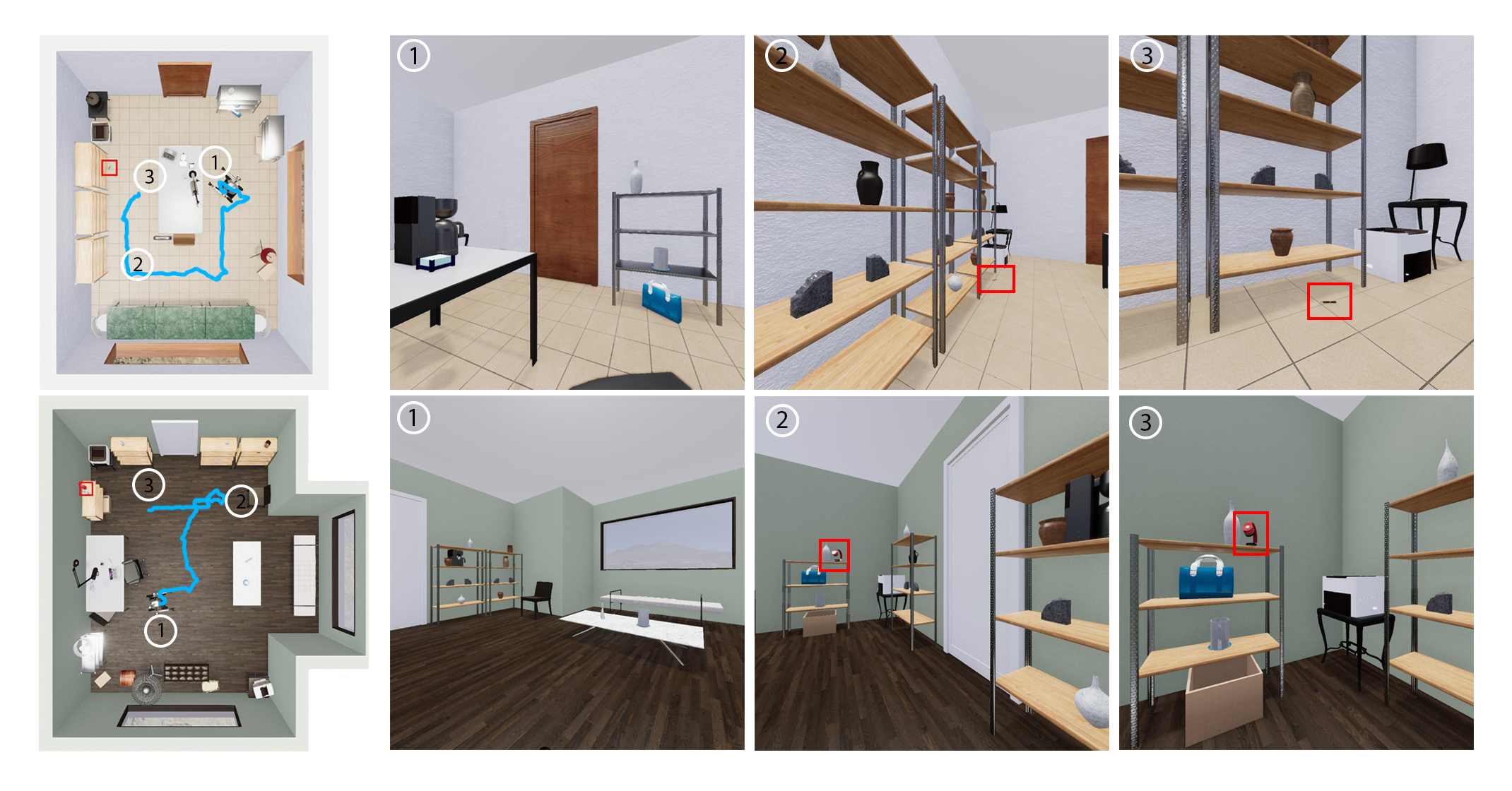}
    \caption{Visualizations of the modular planning based agent. In each case, the top down view shows the agent’s trajectory, while the numbered images show the agent’s egocentric view: 1) at the start, 2) after identifying an object that looks like the fallen object, and 3) after navigating to the object and executing the \textit{found} action. Row 1: fallen object is a pen; Row 2: fallen object is headphones.}
    \label{fig:modular}

\end{figure*}

\subsection{Result Analysis}
\noindent\textbf{Overall results.}
In Table~\ref{tab:main}, we report the overall performance of different agents on 1000 test scenes under the three evaluation metrics. The modular planning-based approach achieves the best results across all metrics but still fails in over half of the scenes. The RL model (PPO) performs poorly on the task, even when adopting an oracle \textit{found}. Surprisingly, we observe that the previous best semantic audio-visual model~\cite{chen2021semantic} (SAVi) also struggles on this task. The reason might be that this model tends to capture the correlations between the scene context and sound. Our dataset tends to eliminate these kinds of shortcuts, because the physical object could fall anywhere in the rooms. The introduction of distractor objects might also pose an addition challenge for the cross-modality attention mechanism in ~\cite{chen2021semantic}. The Decision Transformer achieves a success rate of 17\%, which is comparable to that of the PPO. This result suggests that combining supervised learning from expert data with reinforcement learning or modular planning could be a promising direction for the future. We also notice that the modular SLAM-based framework for object-goal navigation could achieve around 22\% success rate, much lower than that obtained using the audio signal to define the goal location (41\%). This indicates the essential need to incorporate multi-modal physical reasoning capabilities into this kind of modular planning-based algorithm to succeed on this task. 

\begin{table*}[!t]
    \centering
    \caption{Evaluation of cross-scene generalization of different agents.}
   
    \begin{tabular}{l|cccccc}
   \hline
& \multicolumn{3}{c}{Kitchen to Study Room}  &  \multicolumn{3}{c}{Study Room to Kitchen}    \\

Method & Success Rate & SPL & SNA & Success Rate & SPL & SNA \\
 \hline
 PPO  (Oracle found)  &  0.11 & 0.10 & 0.10 & 0.05 & 0.04 & 0.05 \\
 SAVi & 0.20 & 0.11 & 0.09 & 0.19 & 0.14 & 0.11 \\
 Decision TransFormer & 0.07 & 0.06 & 0.06 & 0.08 & 0.06 & 0.07 \\
 Object Navigation  & 0.18 & 0.14 & 0.13 & 0.15 & 0.14 & 0.13 \\
 Modular Planning  & 0.34 & 0.23 & 0.20 & 0.35 & 0.22 & 0.19  \\
 \hline
    \end{tabular}
    \label{tab:cross}
\end{table*}

\vspace{1mm}
\noindent\textbf{What Makes This Task Difficult?}
We also perform an in-depth analysis to understand the challenge of this dataset, using the best modular planning-based approach (Table~\ref{tab:main}). We leveraged the modular design of this method, replacing individual components of the modular framework with an oracle model and measuring the effect on performance. We performed this `oracle swap' with the modules for visual perception (\ie object segmentation), auditory object perception (\ie object category predictions from sound), sound source localization (\ie fallen object location), and the exploration strategy. We first implement two oracle agents with perfect visual perception (GT seg.) and auditory object perception (GT obj.), respectively. These experiments yielded two key observations. First, an oracle visual perception module improved the success rate marginally but significantly reduced the number of steps needed to find the target object. Second, an oracle auditory object category prediction model improved the success rate over 20\%, indicating that one major bottleneck is accurately predicting object categories from audio alone. This could indicate that the trained auditory perception module failed to learn to optimally infer physics from sound. The inability to identify the fallen object category is also a major failure mode for all other baselines. One possible future approach to improve auditory recognition is to learn an object-centric neural implicit function~\cite{gao2021objectfolder} as a generative model to approximate an object's sound based on its physical properties, and to adopt an analysis-by-synthesis approach to invert the object physics from sound. When the two oracle visual and audio modules were combined (GT seg.$+$object), the agent performed well (around 86\% success rate). 

We also implement another oracle agent with additional ground truth 2D position of fallen objects (GT seg.$+$object$+$location) and achieve near-perfect results. The failure cases for this agent were mostly cases where the agent got caught in a narrow aisle and could not get out. These examples indicate that safe exploration in physical rooms is another challenge that must be overcome to fully solve this task. One potential solution for that is to learn a semantic search policy to set waypoints~\cite{min2021film}. We leave this to future work. 

\vspace{2mm}
\noindent\textbf{Qualitative Results.} In Figure~\ref{fig:modular}, we have visualized example trajectories of the modular planning-based agent. The agent first estimates the relative distance and orientation of the fallen object and roughly what category this object might belong to. The planner then explores the environment and builds the map while navigating towards the goal.  Once approaching the surrounding goal area, the visual signal dominates the agent's decisions.  For example, after the agent identifies an object that seems consistent with what it hears, it will adjust its goal location, navigate towards it and execute the \textit{found} action. This process intuitively resembles how humans might perform such a task.  We will provide more demo videos in the supplementary materials. 

\subsection{Evaluation on Cross-Scene Generalization}
\label{sec:cross}
Similar to other embodied AI challenges, we also hope to build autonomous navigation agents that generalize to unseen scenarios. To explicitly test agents' generalization abilities, we define two splits of the training set and test set for cross-scene evaluation. In the first split, the models are trained in the study rooms but tested in the kitchens. In the second split, we train models in the kitchens but test them in the study rooms. The results are reported in Table~\ref{tab:cross}. All models have some degree of cross-scene performance drop. For example, the success rate of the modular planning approach decreased by 6\%. We analyzed some failure cases and found that most of them come from the visual perception module. For example, the Mask-RCNN network sometimes fails to detect the fallen objects lying on a different color carpet, which requires agents to take additional time steps to find it. 

\section{Conclusions}

We introduce a new task of embodied physical sound source localization in 3D virtual environments. To study this problem, we use the ThreeDWorld platform to synthesize a large-scale fallen object audio-visual dataset that includes 8000 instances of 30 physical objects dropping in 64 rooms. We implemented and evaluated several baseline agents as a first step to solving this challenge. Our experiments demonstrate that this task is challenging for current state-of-the-art methods. We further carried out a diagnostic analysis on modular planning-based agents to understand the factors that make this task challenging and to identify future directions for improving task performance. We believe this fallen object dataset will open up new opportunities to integrate physical reasoning abilities from multi-sensory data into embodied agents.  We will release the code and dataset upon paper publication. 

\noindent\textbf{Future Works.} The auditory perception model used by implemented embodied agents falls short of inferring physics from sound, one of the major bottlenecks for solving this new embodied AI challenge on finding fallen objects by integrating asynchronous audio-visual data. In future work, we are interested in learning object-centric neural implicit representations of impact sounds that could encode physical properties and transferring the models learned from this high-fidelity simulation to the real world. 

\vspace{2mm}

\noindent\textbf{Acknowledgement.} This work was supported by MIT-IBM Watson AI Lab and its member company Nexplore, ONR MURI (N00014-13-1-0333), DARPA Machine Common Sense program, ONR (N00014-18-1-2847), NSF Grant BCS 1921501, and MERL.


\clearpage
{\small
\bibliographystyle{ieee_fullname}
\bibliography{egbib}
}

\end{document}



\onecolumn
\appendix

\begin{center}
	{
		\Large{\textbf{Supplementary Materials for ``Finding Fallen  Objects \\ Via Asynchronous Audio-Visual Integration''}}
	}
\end{center}

\vspace{15 pt}

\begin{itemize}[leftmargin=*]
    \item In Section~\ref{sec:dataset}, we provide the list of the object categories and materials used in the fallen objects dataset.
    \item In Section~\ref{sec:implementation}, we provide more training details of the visual and auditory perception models.  
    
\end{itemize}

\section{Dataset}
\label{sec:dataset}

\begin{table}[h]
\centering
\caption{The list of object categories and materials in the \textit{Fallen Objects} dataset.}
\resizebox{0.85\linewidth}{!}{
\begin{tabular}{c|c}
\topline
Object Category & \begin{tabular}[c]{@{}c@{}}candle, calculator, flashlight battery, fork, bookend, spoon, toothbrush, \\ kitchen utensil, cup, coaster, vase, key, bottle cork, toy, \\ bowl, soda can, clothes brush, ipod, box, pepper mill, \\ pepper grinder, bottle, toaster, wineglass, jug, watch, knife, \\ pen, headphone, golf ball, shirt button\end{tabular} \\ 
\midline
Material        & \begin{tabular}[c]{@{}c@{}}plastic\_hard, plastic\_soft\_foam, glass, metal, ceramic, \\ stone, cardboard, wood\_soft, wood\_hard, wood\_medium, \\ fabric, leather, paper, rubber\end{tabular}                                                                                                                                                         \\ 
\bottomline
\end{tabular}
}
\end{table}

\section{Implementation Details}
\label{sec:implementation}

\paragraph{Visual perception networks.} We train a Mask-RCNN using Feature Pyramid Networks with ResNet-50 backbone for detection and instance segmentation. We use a dataset of 60,000 images of 30 categories to train this model. We set batch size as 12 and use SGD as an optimizer. The learning rate is 0.001. The model converges after 70K iterations.

\paragraph{Auditory perception networks.} We use ResNet-18 pre-trained on ImageNet-1K as a backbone which takes as input log Mel-scaled spectrograms extracted from waveforms. Following the backbone, we use three linear layers with a hidden dimension of 1024 to serve as a classifier. We update all parameters of the model during training. We set batch size as 128 and use Adam as an optimizer. The learning rate is 0.0003. The model converges within 200 epochs.
